\documentclass[runningheads]{llncs}
\usepackage{latexsym,multirow,xcolor,caption}
\usepackage[para]{footmisc}
\usepackage[misc]{ifsym}
\usepackage{graphicx}
\begin{document}
\def\datasetName{FICLE}
\def\longName{Factual Inconsistency CLassification with Explanation}

\title{Neural models for Factual Inconsistency Classification with Explanations}
%Tathagata, Mukund, Abhinav, Harshit, Aditya, Manish, Vasudeva.

%Ack: Shashwat, Pratyaksh, Vijay.
%
%\titlerunning{Abbreviated paper title}
% If the paper title is too long for the running head, you can set
% an abbreviated paper title here
%

\author{Tathagata Raha\textsuperscript{1}, Mukund Choudhary\textsuperscript{1}, Abhinav Menon\textsuperscript{1}, Harshit Gupta\textsuperscript{1},\\ K V Aditya Srivatsa\textsuperscript{1}, Manish Gupta\textsuperscript{1,2}%\orcidID{0000-0002-2843-3110}
(\Letter), Vasudeva Varma\textsuperscript{1}\\
\vspace{2pt}
\small {\textsuperscript{1}IIIT-Hyderabad, India; 
\textsuperscript{2}Microsoft, Hyderabad, India}\\
\{tathagata.raha,mukund.choudhary,abhinav.m,harshit.g,k.v.aditya\}@research.iiit.ac.in\\ gmanish@microsoft.com, vv@iiit.ac.in
}
\authorrunning{Raha et al.}
% First names are abbreviated in the running head.
% If there are more than two authors, 'et al.' is used.
%
% \institute{
% \email{lncs@springer.com}\\
% \url{http://www.springer.com/gp/computer-science/lncs} \and
% ABC Institute, Rupert-Karls-University Heidelberg, Heidelberg, Germany\\
% \email{\{abc,lncs\}@uni-heidelberg.de}}
% \author{}
\institute{}
\maketitle              % typeset the header of the contribution
\begin{abstract}
%why is the problem important
Factual consistency is one of the most important requirements when editing high quality documents. It is extremely important for automatic text generation systems like summarization, question answering, dialog modeling, and language modeling. 
%comparison with related work
Still, automated factual inconsistency detection is rather under-studied. Existing work has focused on (a) finding fake news keeping a knowledge base in context, or (b) detecting broad contradiction (as part of natural language inference literature). % or (c) text generation to ensure non-toxic, coherent outputs with high fidelity. 
However, there has been no work on detecting and explaining types of factual inconsistencies in text, without any knowledge base in context.
% and textual entailment. Given a premise and a hypothesis, this work broadly classifies whether the premise entails or contradicts the hypothesis. However, there exists no work to classify the kind of contradiction.
%define problem.
In this paper, we leverage existing work in linguistics to formally define five types of factual inconsistencies. Based on this categorization, we contribute a novel dataset, \datasetName{} (\longName{}), with $\sim$8K samples where each sample consists of two sentences (claim and context) annotated with type and span of inconsistency. When the inconsistency relates to an entity type, it is labeled as well at two levels (coarse and fine-grained).
%our approach
Further, we leverage this dataset to train a pipeline of four neural models to predict inconsistency type with explanations, given a (claim, context) sentence pair. 
Explanations include inconsistent claim fact triple, inconsistent context span, inconsistent claim component, coarse and fine-grained inconsistent entity types.
The proposed system first predicts inconsistent spans from claim and context; and then uses them to predict inconsistency types and inconsistent entity types (when inconsistency is due to entities).
%We find that models perform better when they are trained to predict not just inconsistency span but also the relevant fact triple from the claim; inconsistent span from the context and type of inconsistent entity. 
% across multiple settings: encoder-only vs encoder-decoder, 
% Further, we propose novel neural models for classifying a sample into contradiction types.
%%expt results
We experiment with multiple Transformer-based natural language classification as well as generative models, and find that DeBERTa performs the best. Our proposed methods provide a weighted F1 of $\sim$87\% for inconsistency type classification across the five classes.
%A T5 model trained with this scheme leads to a micro-averaged F1 of 0.80 and ?? for inconsistency type detection and inconsistent entity type detection respectively. 
We make the code and dataset publicly available\footnote{\url{https://github.com/blitzprecision/FICLE}\label{datafootnote}}.
\end{abstract}

\section{Introduction}
\label{sec:intro}

\begin{figure}[!t]
    \centering
    \includegraphics[width=0.6\columnwidth]{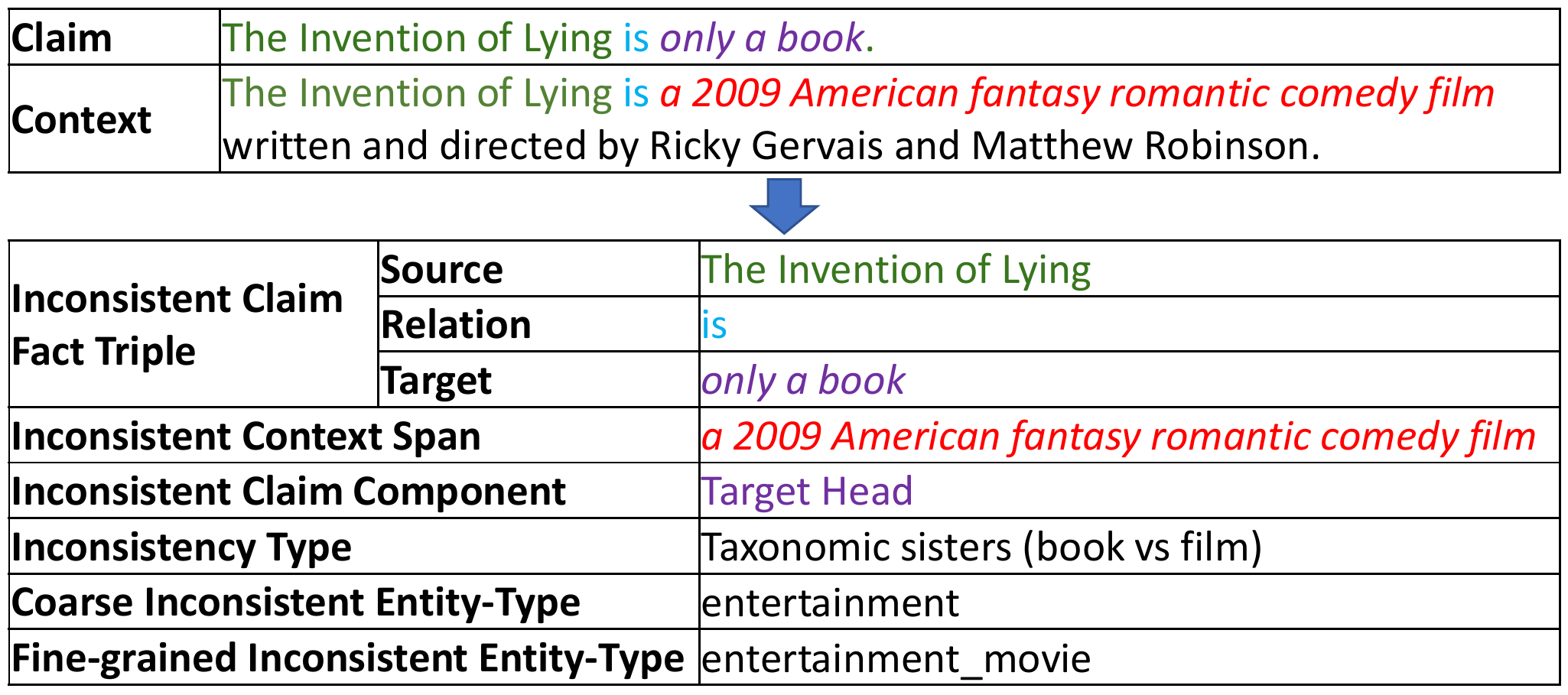}
    \caption{Factual Inconsistency Classification with Explanation (\datasetName{}) Example: Inputs are claim and context. Outputs include inconsistency type and explanation (inconsistent claim fact triple, inconsistent context span, inconsistent claim component, coarse and fine-grained inconsistent entity types).}
    \label{fig:example}
\end{figure}
%\textcolor{red}{Put examples from \url{https://mirandrom.github.io/litreview/2020-03-14-factual-consistency}. Also put examples from other systems like QA/dialog systems. Some examples from our dataset.}

% motivation
%Why is consistency important? Give examples of summarization and dialog based systems by looking at motivation in those papers.
Although Transformer-based natural language generation models have been shown to be state-of-the-art for several applications like summarization, dialogue generation, question answering, table-to-text, and machine translation, they suffer from several drawbacks of which hallucinatory and inconsistent generation is the most critical~\cite{ji2022survey}. %Krysinski et al.~\cite{kryscinski2019neural} and Cao et al.~\cite{cao2018faithful} found that around 30\% of the summaries generated by state-of-the-art abstractive models were factually inconsistent. Similarly, several studies have pointed out semantic inaccuracy as a major problem with current natural language generation models for free-form text generation~\cite{brown2020language}, data-to-text~\cite{duvsek2020evaluating}, question-answering~\cite{longpre2021entity}, dialogue modeling~\cite{mesgar2021improving} etc.
%Cao et al.~\cite{} report that $\sim$30\% of summaries generated by a seq2seq model with attention are factually inconsistent, using a sample of 100 summaries from the Gigaword dataset.
% Why is finegrained contradiction detection important? Why explanations matter?
Factual inconsistencies in generated text can lead to confusion and a lack of clarity, make the text appear unreliable and untrustworthy, and can create a sense of mistrust among readers. It can lead to inaccurate conclusions and interpretations, and diminishes the overall quality of the text. One approach to tackle this problem is to train robust neural language generation models which produce text with high fidelity and less hallucinations~\cite{ji2022survey}. Another approach is to have human annotators post-check the generated text for inconsistencies. Checking all generated output manually is not scalable. Hence, automated factual inconsistency detection and explanations become crucial. 

% Related work drawbacks, drawbacks of other related datasets.
%cite manning paper
%cite fever dataset
Accordingly, there have been several studies in the past which focus on detection of false or fake content. Fake content detection studies~\cite{ciampaglia2015computational,shi2016discriminative,vedula2021face} typically verify facts in claims with respect to an existing knowledge base. However, keeping the knowledge base up-to-date (freshness and completeness) is difficult. Accordingly, there have been other studies in the natural language inference (or textual entailment) community~\cite{bowman2021large,nie2020adversarial,williams-etal-2018-broad} where the broad goal is to predict entailment, contradiction or neither. More than a decade back, De Marneffe et al.~\cite{de2008finding} proposed the problem of fine-grained contradiction detection, but (1) they proposed a tiny dataset with 131 examples, (2) they did not propose any learning method, and (3) they did not attempt explanations like localization of inconsistency spans in claim and context.

% Problem Statement in brief. Example. % What are the challenges?
Hence, in this paper, we propose the novel problem of \underline{f}actual \underline{i}nconsistency \underline{cl}assification with \underline{e}xplanations (\datasetName{}). Given a (claim, context) sentence pair, our goal is to predict inconsistency type and explanation (inconsistent claim fact triple, inconsistent context span, inconsistent claim component, coarse and fine-grained inconsistent entity types). Fig.~\ref{fig:example} shows an example of the \datasetName{} task. Two recent studies are close to our work: e-SNLI~\cite{camburu2018snli} and TaxiNLI~\cite{joshi2020taxinli}. Unlike detailed structured explanation (including inconsistency localization spans in both claim and context) from our proposed system, e-SNLI~\cite{camburu2018snli} contains only an unstructured short sentence as an explanation. Unlike five types of inconsistencies detected along with explanations by our proposed system, TaxiNLI~\cite{joshi2020taxinli} provides a two-level categorization for the NLI task. Thus, TaxiNLI focuses on NLI and not on inconsistencies specifically. Table~\ref{tab:datasetComparison} shows a comparison of our dataset with other closely related datasets.

\begin{table}[!t]
    \centering
    \scriptsize
    \begin{tabular}{|l|c|c|c|c|}
    \hline
Dataset&\#Samples&Explanations&\#Classes&Inconsistency localized?\\
\hline
\hline
Contradiction~\cite{de2008finding}&131&No&10&No\\
\hline
FEVER~\cite{thorne2018fever}&43107&No&1&No\\
\hline
e-SNLI~\cite{camburu2018snli}&189702&Yes&1&Yes\\
\hline
% NILE~\cite{kumar2020nile}&&Yes&&No\\
% \hline
TaxiNLI~\cite{joshi2020taxinli}&3014&No&15&No\\
\hline
LIAR-PLUS~\cite{alhindi2018your}&5669&Yes&3&No\\
\hline
\datasetName{} (Ours)&8055&Yes&5&Yes\\
\hline
    \end{tabular}
    \caption{Comparison of \datasetName{} with other datasets. \#Samples indicates number of contradictory/inconsistent samples (and not the size of full dataset).}
    \label{tab:datasetComparison}
\end{table}

% FICLE dataset.
In this work, based on linguistic theories, we carefully devise a taxonomic categorization with five inconsistency types: simple, gradable, set-based, negation, taxonomic relations. First, we obtain English (claim, context) sentence pairs from the FEVER dataset~\cite{thorne2018fever} which have been labeled as contradiction. We get them manually labeled with inconsistency types and other explanations (as shown in Fig.~\ref{fig:example} by four annotators. Overall, the dataset contains 8055 samples labeled with five inconsistency types, 20 coarse inconsistent entity types and 60 fine-grained inconsistent entity types, whenever applicable. 

% modeling details. % Specific design in models to overcome those challenges
We leverage the contributed dataset to train a pipeline of four neural models to predict inconsistency type with explanations: $M_1$, $M_2$, $M_3$ and $M_4$. Given a (claim, context) sentence pair, $M_1$ predicts the inconsistent subject-relation-target fact triple $\langle S,R,T\rangle$ in the claim and also the inconsistent span in the context. $M_2$ uses $M_1$'s outputs to predict the inconsistency type and the inconsistent component (subject, relation or target) from the claim. $M_3$ uses the inconsistent context-span and inconsistent claim component to predict a coarse inconsistent entity type. $M_4$ leverages both $M_3$'s inputs and outputs to predict fine-grained inconsistent entity type. Overall, the intuition behind this pipeline design is to first predict inconsistent spans from claim and context; and then use them to predict inconsistency types and inconsistent entity types (when inconsistency is due to entities). Fig.~\ref{fig:arch} shows the overall system architecture for \datasetName{}. 

%Results in brief -- overall contributions.
We investigate effectiveness of multiple standard Transformer~\cite{vaswani2017attention}-based natural language understanding (NLU) as well as natural language generation (NLG) models as architectures for models $M_1$, $M_2$, $M_3$ and $M_4$. Specifically, we experiment with models like BERT~\cite{devlin2018bert}, RoBERTa~\cite{liu2019roberta} and DeBERTa~\cite{he2020deberta} which are popular for NLU tasks. We also experiment with T5~\cite{raffel2020exploring} and BART~\cite{lewis2020bart} which are popular in the NLG community. DeBERTa seemed to outperform other models for most of the sub-tasks. Our results show that while inconsistency type classification is relatively easy, accurately detecting context span is still challenging.

Overall, in this work, we make the following main contributions.
% \begin{itemize}
    (1) We propose a novel problem of factual inconsistency detection with explanations given a (claim, context) sentence pair. 
    (2) We contribute a novel dataset, \datasetName{}, manually annotated with inconsistency type and five other forms of explanations. We make the dataset publicly available\footref{datafootnote}.
    (3) We experiment with standard Transformer-based NLU and NLG models and propose a baseline pipeline for the \datasetName{} task.
    (4) Our proposed pipeline provides a weighted F1 of $\sim$87\% for inconsistency type classification; weighted F1 of $\sim$86\% and $\sim$76\% for coarse (20-class) and fine-grained (60-class) inconsistent entity-type prediction respectively; and an IoU of $\sim$94\% and $\sim$65\% for claim and context span detection respectively.
% \end{itemize}

%Paper organization.
% The remainder of the paper is organized as follows. We discuss related work on natural language inference and factual inconsistency in Section~\ref{sec:relWork}. We propose the five-class inconsistency type taxonomy in Section~\ref{sec:taxonomy}. We discuss details of dataset curation, pre-processing and analysis in Section~\ref{sec:dataset}. We present our four-model pipeline for the \datasetName{} task in Section~\ref{sec:approach}. We discuss details of our experiments and insights in Section~\ref{sec:experiments}. Finally, we conclude with a brief summary in Section~\ref{sec:conclusion}.

\section{Related Work}
\label{sec:relWork}
\noindent\textbf{Factual Inconsistency in Natural Language Generations}: 
Popular natural language generation models have been found to generate hallucinatory and inconsistent text~\cite{ji2022survey}. Krysinski et al.~\cite{kryscinski2019neural} and Cao et al.~\cite{cao2018faithful} found that around 30\% of the summaries generated by state-of-the-art abstractive models were factually inconsistent. There are other summarization studies also which report factual inconsistency of generated summaries~\cite{mao2020constrained,maynez2020faithfulness,nan2021entity,wang2020asking,zhang2021fine,zhu2021enhancing}.
% Populate this section based on \url{https://mirandrom.github.io/litreview/2020-03-14-factual-consistency} and abstracts of other referenced papers.
Similarly, several studies have pointed out semantic inaccuracy as a major problem with current natural language generation models for free-form text generation~\cite{brown2020language}, data-to-text~\cite{duvsek2020evaluating}, question-answering~\cite{longpre2021entity}, dialogue modeling~\cite{honovich2021q,mesgar2021improving}, machine translation~\cite{zhou2021detecting}, and news generation~\cite{zellers2019defending}. Several statistical (like PARENT) and model-based metrics have been proposed to quantify the level of hallucination. Multiple data-related methods and modeling and inference methods have been proposed for mitigating hallucination~\cite{ji2022survey}, but their effectiveness is still limited. Hence, automated factual inconsistency detection is critical.

\noindent\textbf{Natural Language Inference}: Natural language inference (NLI) is the task of determining whether a hypothesis is true (entailment), false (contradiction), or undetermined (neutral) given a premise. NLI is a fundamental problem in natural language understanding and has many applications such as question answering, information extraction, and text summarization. Approaches used for NLI include earlier symbolic and statistical approaches to more recent deep learning approaches~\cite{bowman2015large}. There are several datasets and benchmarks for evaluating NLI models, such as the Stanford Natural Language Inference (SNLI) Corpus~\cite{bowman2021large}, the Multi-Genre Natural Language Inference (MultiNLI) Corpus~\cite{williams-etal-2018-broad} and  Adversarial NLI~\cite{nie2020adversarial}. FEVER~\cite{thorne2018fever} is another dataset on a related problem of fact verification.

Recently there has been work on providing explanations along with the classification label for NLI. e-SNLI~\cite{camburu2018snli} provides a one-sentence explanation aiming to answer the question: ``Why is a pair of sentences in a relation of entailment, neutrality, or contradiction?'' Annotators were also asked to highlight the words that they considered essential for the label. NILE~\cite{kumar2020nile} is a two stage model built on e-SNLI which first generates candidate explanations and then processes explanations to infer the task label. Thorne et al.~\cite{thorne2019generating} evaluate LIME~\cite{ribeiro2016should} and Anchor explanations~\cite{ribeiro2018anchors} to predict token annotations that explain the entailment relation in e-SNLI. LIAR-PLUS~\cite{alhindi2018your} contains political statements labeled as pants-fire, false, mostly-false, half-true, mostly-true, and true. The context and explanation is combined into a ``extracted justification'' paragraph in this dataset. Atanasova et al.~\cite{atanasova2020generating} experiment with LIAR-PLUS dataset and find that jointly generating justification and predicting the class label together leads to best results. 

There has also been work on detailed categorization beyond just the two classes: contradiction and entailment. Contradiction~\cite{de2008finding} is a tiny dataset with only 131 examples that provides a taxonomy of 10 contradiction types. Recently, TaxiNLI~\cite{joshi2020taxinli} dataset has been proposed with 15 classes for detailed categorization with the entailment and not the contradiction category. 
Continuing this line of work, in this paper, we contribute a new dataset, \datasetName{}, which associates every (claim, context) sentence pair with (1) an inconsistency type (out of five) and (2) detailed explanations (inconsistent span in claim and context, inconsistent claim component, coarse and fine-grained inconsistent entity types).

% \subsection{Contradiction Detection}
% ~\cite{de2008finding,li2018computational,zadrozny2017towards,ellis1997emotion,kalouli2017correcting,tsytsarau2011towards}

\section{Inconsistency Type Classification}
\label{sec:taxonomy}

Factual inconsistencies in text can occur because of a number of different sentence constructions, some overt and others that are complex to discover even manually. We design a taxonomy of five inconsistency types following non-synonymous lexical relations classified by Saeed~\cite[p.~66--70]{saeed2011semantics}. The book mentions the following kinds of antonyms: simple, gradable, reverses, converses and taxonomic sisters. To this taxonomy, we added two extra categories, negation and set-based, to capture the \datasetName{}'s complexity. Also, we expanded the definition of taxonomic sisters to more relations, and hence rename it to taxonomic relations. Further, since we did not find many examples of reverses and converses in our dataset, we merged them with the simple inconsistency category.
%http://coursdelinguistique.free.fr/Ressources/Semantics.pdf
Overall, our \datasetName{} dataset contains these five different inconsistency types.

    \begin{itemize}
        \item Simple: A simple contradiction is a direct contradiction, where the negative of one implies the positive of the other in a pair like \emph{pass vs. fail}. This also includes actions/ processes that can be reversed or have a reverse direction, like \emph{come vs. go} and \emph{fill vs. empty}. Pairs with alternate viewpoints like \emph{employer vs. employee} and \emph{above vs. below} are also included in this category.
        \item Gradable: Gradable contradictions include adjectival and relative contradictions, where the positive of one, does not imply the negative of other in a pair like \emph{hot vs.  cold}, \emph{least vs. most}, or periods of time etc.
        \item Taxonomic relations: We include three kinds of relations in this type: (a) Pairs at the same taxonomic level in the language like \emph{red vs. blue} which are placed parallel to each other under the English color adjectives hierarchy. (b) When a pair has a more general word (\emph{hypernym}) and another more specific word which includes the meaning of the first word in the pair (\emph{hyponym}) like \emph{giraffe} (hypo) vs. \emph{animal} (hyper). (c) Pairs with a part-whole relation like \emph{nose vs. face} and \emph{button vs shirt}.
        \item Negation: This includes inconsistencies arising out of presence of explicit negation morphemes (e.g. \textit{not},  \textit{except}) or a finite verb negating an action  (e.g. \textit{fail to do X}, \textit{incapable of X-ing}) etc.
        \item Set-based: This includes inconsistent examples where an object contrasts with a list that it is not a part of (e.g. \textit{cat} vs. \textit{bee, ant, wasp}).
\end{itemize}

\section{The \datasetName{} Dataset}
\label{sec:dataset}
\begin{table*}[!b]
    \centering
    \scriptsize
    \begin{tabular}{|p{0.6\textwidth}|p{0.2\textwidth}|p{0.2\textwidth}|}
    \hline
     Inconsistent Claim&Inconsistent Context Span&Inconsistent Claim Component\\
          \hline
               \hline
     \textbf{Prime Minister Swami Vivekananda} \textit{enthusiastically hoisted} \underline{the Indian flag}.&Narendra Modi&Subject-Head\\
                    \hline
\textbf{President Narendra Modi} \textit{enthusiastically hoisted} \underline{the Indian flag}.&Prime Minister&Subject-Modifier\\
               \hline
     \textbf{Prime Minister Narendra Modi} \textit{enthusiastically lowered} \underline{the Indian flag}.&hoisted&Relation-Head\\
                    \hline
\textbf{Prime Minister Narendra Modi} \textit{halfheartedly hoisted} \underline{the Indian flag}.&enthusiastically&Relation-Modifier\\
               \hline
     \textbf{Prime Minister Narendra Modi} \textit{enthusiastically hoisted} \underline{the Indian culture}.&flag&Target-Head\\
                    \hline
\textbf{Prime Minister Narendra Modi} \textit{enthusiastically hoisted} \underline{the American flag}.&Indian&Target-Modifier\\
 \hline
    \end{tabular}
    \caption{Inconsistent Claim Fact Triple, Context Span and Claim Component examples for the context sentence ``Prime Minister Narendra Modi enthusiastically hoisted the Indian flag.'' Subject, relation and target in the claim are shown in bold, italics and underline respectively.}
    \label{tab:syntacticExamples}
\end{table*}

\begin{table*}[!t]
    \centering
    \scriptsize
    \begin{tabular}{|p{0.22\textwidth}|p{0.32\textwidth}|p{0.12\textwidth}|p{0.15\textwidth}|p{0.15\textwidth}|}
    \hline
    Claim&Context&Incon-sistency Type&Coarse Inconsistent Entity Type    & Fine-grained Inconsistent Entity Type \\
    \hline
    \hline
Kong: Skull Island \textbf{is not a} reboot.&The film \textbf{is a} reboot of the King Kong franchise and serves as the second film in Legendary's MonsterVerse .&Negation&enter-tainment&brand\\
\hline
The Royal Tenenbaums only stars \textbf{Emma Stone}.&The film stars \textbf{Danny Glover, Gene Hackman, Anjelica Huston, Bill Murray, Gwyneth Paltrow, Ben Stiller, Luke Wilson, and Owen Wilson}.&Set Based&name&musician\\
\hline
Lindsay Lohan began her career as \textbf{an adult fashion model}.&Lohan began her career as \textbf{a child fashion model} when she was three, and was later featured on the soap opera Another World for a year when she was 10 .&Simple&time&age\\
\hline
Karl Malone played the \textbf{shooting guard position}.&He is considered one of the best \textbf{power forwards} in NBA history .&Taxonomic Relation&profession&sport\\
\hline
The Divergent Series: Insurgent is based on the \textbf{third} book in the Divergent trilogy.&The Divergent Series : Insurgent is a 2015 American science fiction action film directed by Robert Schwentke, based on Insurgent, the \textbf{second} book in the Divergent trilogy by Veronica Roth.&Gradable&quantity&ordinal\\
\hline
    \end{tabular}
    \caption{Inconsistency Type and Coarse/Fine-grained Inconsistent Entity Type examples. Inconsistent spans are marked in bold in both claim as well as context.}
    \label{tab:semanticExamples}
\end{table*}
% quick 2-liner intro
\subsection{Dataset Curation and Pre-processing}
Our \datasetName{} dataset is derived from the FEVER dataset~\cite{thorne2018fever} using the following processing steps. FEVER (Fact Extraction and VERification) consists of 185,445 claims generated by altering sentences extracted from Wikipedia and subsequently verified without knowledge of the sentence they were derived from. Every sample in the FEVER dataset contains the claim sentence, evidence (or context) sentence from a Wikipedia URL, a type label (`supports', `refutes' or `not enough info'). Out of these, we leverage only the samples with `refutes' label to build our dataset.

%The purpose of the FEVER challenge is to evaluate the ability of a system to verify information using evidence from Wikipedia. 
%For the FEVER dataset, we were provided with the Wikipedia data dump and claim sentences. For every claim sentence, the task was to extract textual evidence from the Wikipedia dump to label the claim as Supported, Refuted given the evidence or NotEnoughInfo. 
%We have leveraged the claims with "Refuted" label to build our dataset. For every "Refuted" claim, we have extracted the evidence sentences from the Wikipedia dump which form the context sentences in our dataset.

% \subsection{Building the dataset}
We propose a linguistically enriched dataset to help detect inconsistencies and explain them. To this end, the broad requirements are to locate where an inconsistency is present between a claim and a context, and to have a classification scheme for better explainability. %Expanding on Marneffe's taxonomy of capturing ``incompability'' in text~\cite{de2008finding} we aim to explain as much as we can using generalized but fine-grained: lexical and other novel linguistic taxonomies which can be partially mapped back to classes like Numeric, Factive etc. and also magnifies into all-inclusive tags like "World Knowledge".

\subsection{Annotation Details}
To support detailed inconsistency explanations, we perform comprehensive annotations for each sample in the \datasetName{} dataset. The annotations were done in two iterations. The first iteration focused on ``syntactic oriented'' annotations while the second iteration focused on ``semantic oriented'' annotations. The annotations were performed using the Label Studio annotation tool\footnote{\url{https://labelstud.io/}} by a group of four annotators (two of which are also authors). The annotators are well versed in English and are Computer Science Bachelors students with a specialization in computational linguistics, in the age group of 20--22 years. Detailed annotation guidelines are in annotationGuidelines.pdf here\footref{datafootnote}.

\noindent\textbf{Syntactic Oriented Annotations:} In this annotation stage, the judges labeled the following syntactic fields per sample. Table~\ref{tab:syntacticExamples} shows examples of each of these fields.
% \begin{itemize}
    (1) Inconsistent Claim Fact Triple: A claim can contain multiple facts. The annotators identified the fact that is inconsistent with the context. Further, the annotators labeled the span of source (S), relation (R) and target (T) within the claim fact. Sometimes, e.g., in case of an intransitive verb, the target was empty. Further, for each of the S, R and T, the annotators also labeled head and modifier separately. The head indicates the main noun (for S and T) or the verb phrase (for R) while the modifier is phrase that serves to modify the meaning of the noun or the verb.
    (2) Inconsistent Context Span: A span marked in the context sentence which is inconsistent with the claim.
    (3) Inconsistent Claim Component: This can take six possible values depending on the part of the claim fact triple that is inconsistent with the context: Subject-Head, Subject-Modifier, Relation-Head, Relation-Modifier, Target-Head, Target-Modifier.
% \end{itemize}

    % \item \textbf{Mismatch Location}: We explicitly annotate which type of Span is mismatching (among S, R, and T) and if it is the head of the span or the modifier - \{ \underline{Source-Head}, \underline{Source-Modifier}, \underline{Relation-Head}, \underline{Relation-Modifier}, \underline{Target-Head}, and \underline{Target} \underline{-Modifier} \}

\noindent\textbf{Semantic Oriented Annotations:}
In this annotation stage, the annotators labeled the following semantic fields per sample. Table~\ref{tab:semanticExamples} shows examples of each of these fields.
% \begin{itemize}
(1) Inconsistency Type: Each sample is annotated with one of the five inconsistency types as discussed in Section~\ref{sec:taxonomy}.
(2) Coarse Inconsistent Entity Type: When the inconsistency is because of an entity, the annotator also labeled one of the 20 coarse types for the entity causing the inconsistency. The types are action, animal, entertainment, gender, geography, identity, material, name, nationality, organization, others, politics, profession, quantity, reality, relationship, sentiment, sport, technology and time.
(3) Fine-grained Inconsistent Entity Type: Further, when the inconsistency is because of an entity, the annotator also labeled one of the 60 fine-grained types for the entity causing the inconsistency.
% \end{itemize}

For inconsistency entity type detection, the annotations were performed in two iterations. In the first iteration, the annotators were allowed to annotate the categories (both at coarse and fine-grained level) freely without any limited category set. This was performed on 500 samples. The annotators then discussed and de-duplicated the category names. Some rare categories were merged with frequent ones. This led to a list of 20 coarse and 60 fine-grained entity types (including ``others''). In the second iteration, annotators were asked to choose one of these categories.
We measured inter-annotator agreement on 500 samples. For source, relation, target and inconsistent context spans, the intersection over union (IoU) was found to be 0.91, 0.83, 0.85 and 0.76 respectively. Further, the Kappa score was found to be 0.78, 0.71 and 0.67 for the inconsistency type, coarse inconsistent entity type and fine-grained inconsistent entity type respectively.

%We mark spans of linguistic triples of two focus (part of the claim that is mismatching with the context) entities, the relation between them, and that of the mismatch in the context. We also mark the linguistic kind of mismatch and the taxonomic kind of entity where applicable.

\begin{figure}
    \centering
\begin{minipage}{0.58\textwidth}
    \includegraphics[width=0.9\columnwidth]{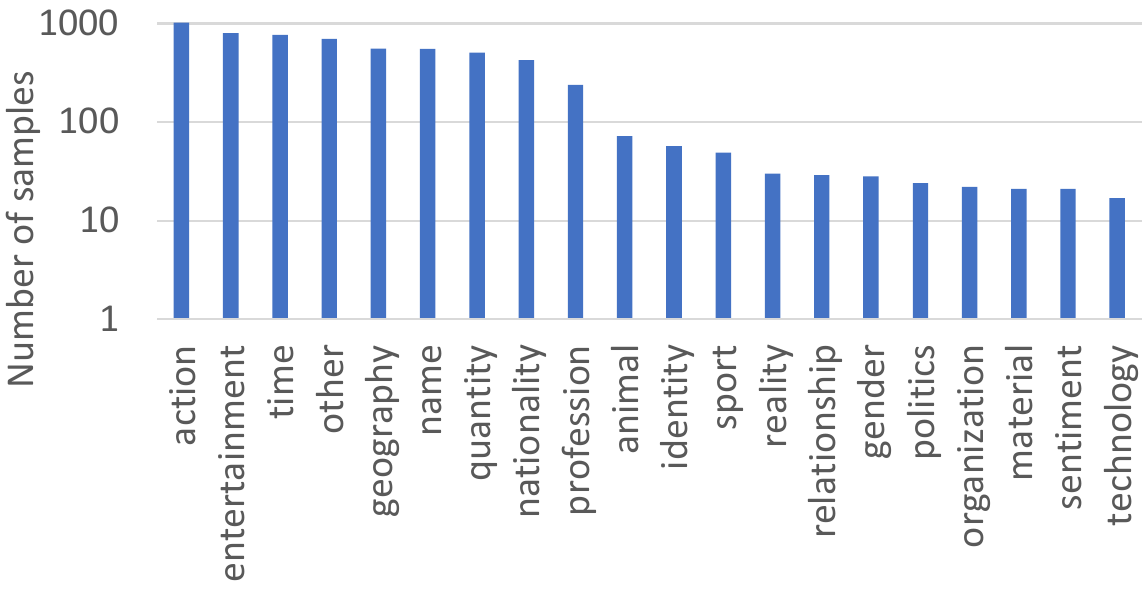}
    \caption{Distribution of coarse inconsistent entity types in \datasetName{}.}
    \label{fig:entityDist}
\end{minipage}
\hfill
\begin{minipage}{0.38\textwidth}
\scriptsize
    \begin{tabular}{|l|p{0.8in}|c|c|c|}
\hline
&&Min&Avg&Max\\
\hline
\hline
&Claim&3&8.04&31\\
\hline
&Context&5&30.73&138\\
\hline
\multirow{3}{*}{\rotatebox{90}{\parbox{0.8cm}{Incon. Claim}}}&Source&1&2.29&9\\
\cline{2-5}
&Relation&1&2.17&18\\
\cline{2-5}
&Target&0&3.39&21\\
\hline
&Incon. Context-Span&1&5.11&94\\
\hline
    \end{tabular}
    \captionof{table}{Minimum, average, and maximum size (words) of various fields averaged across samples in \datasetName{} dataset.}
    \label{tab:datasetStats}
\end{minipage}
\end{figure}

% \begin{table}
%     \centering
%     \scriptsize

% \end{table}

% The dataset showed that a lot of mismatching entities were under the Taxonomic Sisters Mismatch Type label, thus necessitating a parallel a fine-grained classification for entity pair mismatches. Following, we mark the kind of entity that is mismatching by designing a tagset inspired by Glockner's experiments ~\cite{Glockner_2018} with lexical replacements to test NLI systems and Panenghat's ~\cite{paul-panenghat-etal-2020-towards} delexicalising tests with synsets to debais NLI datasets against specific lexical items. We provide this abstract lexical information for entities in form of coarse class tags like - \emph{entertainment}, \emph{identity}, \emph{sport} etc. and narrower ones like - \emph{country, state} etc. for coarse class tag \emph{geography} and \emph{year, month} etc. for \emph{time} among others.

% \href{https://docs.google.com/spreadsheets/d/1JH50-fHuwF3OX-H5WAfiLRjEepxlpknD5uPzRqQwrEs/edit?usp=sharing}{Table} \ref{tab:datasetExample} (linked) shows an annotated example with the possible tagset for every column. 

% \textcolor{red}{Put up a table with an example for every type. Type, claim, context, S,R, T, mismatch entity type, span.}

% mention that the detailed guidelines are mentioned in appendix
\subsection{\datasetName{} Dataset Statistics}
The \datasetName{} dataset consists of 8055 samples in English with five inconsistency types. The distribution across the five types is as follows: Taxonomic Relations (4842), Negation (1630), Set Based (642), Gradable (526) and Simple (415). There are six possible inconsistent claim components with distribution as follows: Target-Head (3960), Target-Modifier (1529), Relation-Head (951), Relation-Modifier (1534), Source-Head (45), Source-Modifier (36). The dataset contains 20 coarse inconsistent entity types as shown in Fig.~\ref{fig:entityDist}. Further, these are sub-divided into 60 fine-grained entity types. Table~\ref{tab:datasetStats} shows average sizes of various fields averaged across samples in the dataset.  The dataset was divided into train, valid and test splits in the ratio of 80:10:10.

%\href{https://docs.google.com/spreadsheets/d/1JH50-fHuwF3OX-H5WAfiLRjEepxlpknD5uPzRqQwrEs/edit#gid=1950890472}{Table}~\ref{tab:datasetStats} (full table linked) shows basic statistics of the \textsc{Mismatch?} dataset. 
%The dataset was annotated by four annotators and we have a total of 10985 samples. (More to be added after discussion with Mukund and Manish sir). (TOFILL do we explain folded versions?)

%distribution across mismatch entity types like Fig 2 at https://arxiv.org/pdf/2104.08836v3.pdf

%min/Avg/median/max size of claim, context, span, subject, predicate, object. 

% \subsection{Quality checks}
% \begin{enumerate}
%     \item S, R, T, and M should be exactly tagged once in an annotated sample
%     \item The S, R, and T spans should be marked on the Claim and M on the Context sentence
%     \item Check the consistently erroneous cases where a model which predicts S, R, T etc. very well, has not been able to predict accurately.
%     \item Check if Mismatch location is marked as Target-Head or Target-Modifier when there is no Target in the sentence etc
%     \item For annotators that have more cases marked as bad tagging ones as compared to others, we do more stringent random checks.
% \end{enumerate}

\section{Neural Methods for Factual Inconsistency Classification with Explanations}
\label{sec:approach}
\begin{figure*}[!t]
    \centering
    \includegraphics[width=0.8\textwidth]{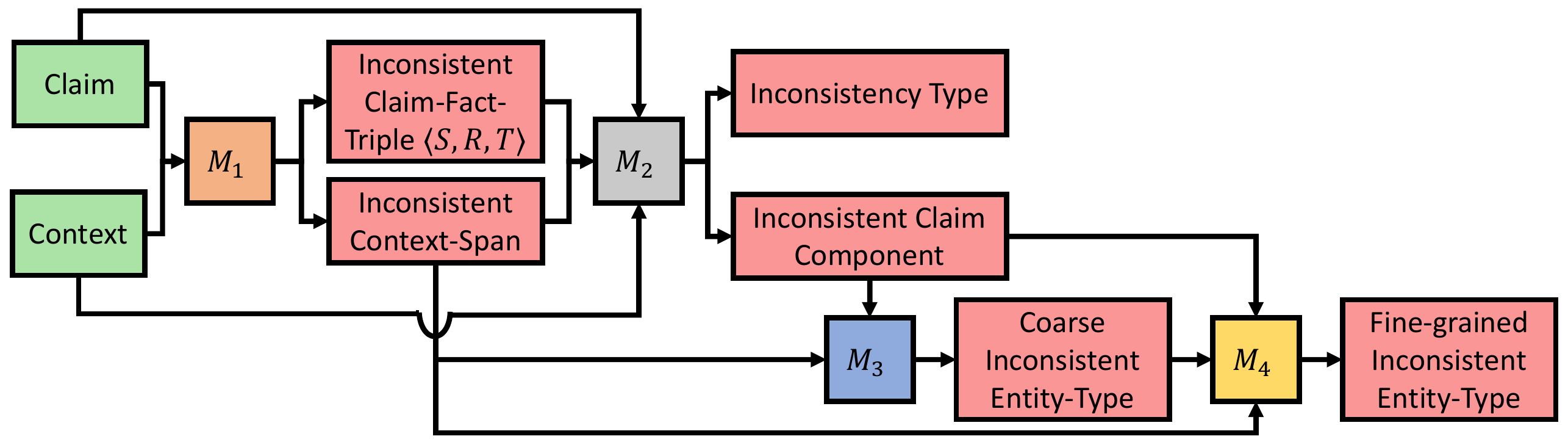}
    \caption{\datasetName{}: System Architecture}
    \label{fig:arch}
\end{figure*}

We leverage the \datasetName{} dataset to train models for factual inconsistency classification with explanations. Specifically, given the claim and context sentence, our system does predictions in the following stages: (A) Predict Inconsistent Claim Fact Triple (S,R,T) and  Inconsistent Context Span, (B) Predict Inconsistency Type and Inconsistent Claim Component, (C) Predict Coarse  and Fine-grained Inconsistent Entity Type. Overall, the system architecture consists of a pipeline of four neural models to predict inconsistency type with explanations: $M_1$, $M_2$, $M_3$ and $M_4$, and is illustrated in Fig.~\ref{fig:arch}. We discuss details of the three stages and the pipeline in this section. 

\noindent\textbf{Model Architectures}
We experiment with five pretrained models of which two are natural language generation (NLG) models. Specifically, we finetune Transformer~\cite{vaswani2017attention} encoder based models like BERT~\cite{devlin2018bert}, RoBERTa~\cite{liu2019roberta} and DeBERTa~\cite{he2020deberta}. We also use two NLG models: BART~\cite{lewis2020bart} and T5~\cite{raffel2020exploring} which are popular in the NLG community.

BERT (Bidirectional Encoder Representations from Transformers)~\cite{devlin2018bert} essentially is a transformer encoder with 12 layers, 12 attention heads and 768 dimensions. We used the pre-trained model which has been trained on Books Corpus and Wikipedia using the MLM (masked language model) and the next sentence prediction (NSP) loss functions. RoBERTa~\cite{liu2019roberta} is a robustly optimized method for pretraining natural language processing (NLP) systems that improves on BERT. RoBERTa was trained with 160GB of text, trained for larger number of iterations up to 500K with batch sizes of 8K and a larger byte-pair encoding (BPE) vocabulary of 50K subword units, without NSP loss. DeBERTa~\cite{he2020deberta} is trained using a special attention mechanism where content and position embeddings are disentangled. It also has an enhanced mask decoder which leverages absolute word positions effectively. BART~\cite{lewis2020bart} is a denoising autoencoder for pretraining sequence-to-sequence models. BART is trained by (1) corrupting text with an arbitrary noising function, and (2) learning a model to reconstruct the original text. T5~\cite{raffel2020exploring} is also a Transformer encoder-decoder model pretrained on Colossal Clean Crawled Corpus, and models all NLP tasks in generative form. 

When encoding input or output for these models, we prepend various semantic units using special tokens like $\langle$claim$\rangle$, $\langle$context$\rangle$,  $\langle$source$\rangle$, $\langle$relation$\rangle$, $\langle$target$\rangle$, $\langle$contextSpan$\rangle$, $\langle$claimComponent$\rangle$, $\langle$type$\rangle$, $\langle$coarseEntityType$\rangle$ and $\langle$fineEntityType$\rangle$. NLG models (BART and T5) generate the inconsistency type and all explanations, and are trained using cross entropy loss. For NLU models (BERT, RoBERTa, DeBERTa), we prepend input with a [CLS] token and use its semantic representation from the last layer with a dense layer to predict inconsistency type, inconsistent claim component, and entity types with categorical cross entropy loss. With NLU models, source, relation, target, and context span are predicted using start and end token classifiers (using cross entropy loss) as usually done in the question answering literature~\cite{devlin2018bert}. 

\noindent\textbf{Stage A: Predict Inconsistent Spans}
In this stage, we first train models to predict source, relation and target by passing the claim sentence as input to the models. Further, to predict inconsistent context span, we experiment with four different methods as follows.
% \begin{itemize}
    (1) Structure-ignorant: The input is claim and context sentence. The aim is to directly predict inconsistent context span ignoring the ``source, relation, target'' structure of the claim.
    (2) Two-step: First step takes claim and context sentences as input, and predicts source, relation and target (SRT). Second step augments source, relation and target to the input along with claim and context, and predicts the inconsistent context span.
    (3) Multi-task: The input is claim and context sentence. The goal is to jointly predict source, relation, target and inconsistent context span.  
    (4) Oracle-structure: The input is claim and context sentence, and ground truth (source, relation and target). These are all used together to predict inconsistent context span.
% \end{itemize}

\noindent\textbf{Stage B: Predict Inconsistency Type and Claim Component}

This stage assumes that (1) SRT from claim and (2) inconsistent context span have already been predicted. Thus, in this stage, the input is claim, context, predicted SRT and predicted inconsistent context span. Using these inputs, to predict inconsistency type and inconsistent claim component, we experiment with three different methods as follows.
% \begin{itemize}
    (1) Individual: Predict inconsistency type and inconsistent claim component separately.
    (2) Two-step: First step predicts inconsistent claim component. Second step augments the predicted inconsistent claim component to the input, and predicts inconsistency type.
    (3) Multi-task: Jointly predict inconsistency type and inconsistent claim component in a multi-task learning setup.
% \end{itemize}

\noindent\textbf{Stage C: Predict Inconsistent Entity Types}

To find inconsistent entity types, we build several models each of which take two main inputs: inconsistent context span and the span from the claim corresponding to the inconsistent claim component. We experiment with the following different models.
% \begin{itemize}
    (1) Individual: Predict coarse and fine-grained inconsistent entity type separately.
    (2) Two-step: First step predicts coarse inconsistent entity type. Second step augments the predicted coarse inconsistent entity type to the input, and predicts fine-grained type.
% \end{itemize}

Further, we also attempt to leverage semantics from entity class names. Hence, we use the NLU models (BERT, RoBERTa, DeBERTa) to obtain embeddings for entity class names, and train NLU models to predict the class name which is most similar to semantic representation (of the [CLS] token) of the input. We use cosine embedding loss to train these models. Specifically, using class (i.e., entity type) embeddings, we train the following models. Note that we cannot train NLG models using class embeddings; thus we perform this experiment using NLU models only.
% \begin{itemize}
    (1) Individual Embedding: Predict coarse and fine-grained inconsistent entity type separately using entity type embeddings.
    (2) Two-step Embedding: First step predicts coarse inconsistent entity type using class embeddings. Second step augments the predicted coarse inconsistent entity type to the input, and predicts fine-grained type using class embeddings.
    (3) Two-step Mix: First step predicts coarse inconsistent entity type using class embeddings. Second step augments the predicted coarse inconsistent entity type to the input, and predicts fine-grained type using typical multi-class classification without class embeddings.
% \end{itemize}

After experimenting with various model choices for the three stages described in this section, we find that the configuration described in Fig.~\ref{fig:arch} provides best results. We also attempted other designs like (1) predicting all outputs (inconsistency type and all explanations) jointly as a 6-task setting using just claim and context as input, (2) identifying claim component only as S, R or T rather than heads versus modifiers. However, these alternate designs did not lead to better results.

\section{Experiments and Results}
\label{sec:experiments}
% \subsection{Metrics}
For prediction of spans like source, relation, target, and inconsistent context span, we use exact match (EM) and intersection over union (IoU) metrics. EM is a number from 0 to 1 that specifies the amount of overlap between the predicted and ground truth span in terms of tokens. If the characters of the model's prediction exactly match the characters of ground truth span, EM = 1, otherwise EM = 0. Similarly, IoU measures intersection over union in terms of tokens. For classification tasks like inconsistency type prediction as well as coarse and fine-grained inconsistent entity type prediction, we use metrics like accuracy and weighted F1. 

% \subsection{Main Results}

Since factual inconsistency classification is a novel task, there are no existing baseline methods to compare with. 

\noindent\textbf{Source, Relation, Target and Inconsistent Context Span Prediction}: Table~\ref{tab:SRT} shows results for source, relation and target prediction from claim sentences. The table shows that T5 works best except for prediction of relation and target using the exact match metric. Further, Table~\ref{tab:contextSpan} shows that surprisingly structure ignorant method is slightly better than the two-step method. Oracle method with DeBERTa expectedly is the best. NLG models (BART and T5) perform much worse compared to NLU models for context span prediction. Lastly, we show results of jointly predicting source, relation, target and inconsistent context span in Table~\ref{tab:jointSRTContextSpan}. The table shows while T5 and BART are better at predicting source, relation and target, DeBERTa is a clear winner in predicting the inconsistent context span.

\begin{table}[!t]
    \centering
    \scriptsize
    \begin{tabular}{|l|c|c|c|c|c|c|}
    \hline
Model&\multicolumn{3}{c|}{Exact Match}&\multicolumn{3}{c|}{IoU}\\
\hline
&Source&Relation&Target&Source&Relation&Target\\
\hline
\hline
BERT&0.919&0.840&\textbf{0.877}&0.934&0.876&\textbf{0.895}\\
\hline
RoBERTa&0.921&\textbf{0.865}&0.871&0.936&0.883&0.885\\
\hline
DeBERTa&0.918&0.857&0.864&0.932&0.874&0.893\\
\hline
BART&0.981&0.786&0.741&0.986&0.873&0.842\\
\hline
T5&\textbf{0.983}&0.816&0.765&\textbf{0.988}&\textbf{0.945}&0.894\\
\hline
    \end{tabular}
    \caption{Source, Relation and Target Prediction from Claim Sentence}
    \label{tab:SRT}
\end{table}

\begin{table}[!t]
    \centering
        \scriptsize
    \begin{tabular}{|l|p{0.5in}|p{0.4in}|p{0.5in}|p{0.5in}|p{0.4in}|p{0.5in}|}
\hline
Model&\multicolumn{3}{c|}{Exact Match}&\multicolumn{3}{c|}{IoU}\\
\hline
&Structure-ignorant&Two-step&Oracle-structure&Structure-ignorant&Two-step&Oracle-structure\\
\hline
\hline
BERT&0.483&0.499&0.519&0.561&0.541&0.589\\
\hline
RoBERTa&0.542&0.534&0.545&0.589&0.584&0.632\\
\hline
DeBERTa&0.538&0.540&\textbf{0.569}&0.591&0.587&\textbf{0.637}\\
\hline
BART&0.427&0.292&0.361&0.533&0.404&0.486\\
\hline
T5&0.396&0.301&0.352&0.517&0.416&0.499\\
\hline
    \end{tabular}
    \caption{Inconsistent Context Span Prediction}
    \label{tab:contextSpan}
\end{table}

\begin{table}[!t]
    \centering
        \scriptsize
        \begin{tabular}{|l|c|c|c|c|c|c|c|c|}
\hline
Model&\multicolumn{4}{c|}{Exact Match}&\multicolumn{4}{c|}{IoU}\\
\hline
&Source&Relation&Target&Context Span&Source&Relation&Target&Context Span\\
\hline
\hline
BERT&0.769&0.665&0.752&0.524&0.801&0.708&0.804&0.566\\
\hline
RoBERTa&0.759&0.686&0.780&0.572&0.828&0.745&0.836&0.617\\
\hline
DeBERTa&0.788&0.704&0.819&\textbf{0.604}&0.843&0.768&0.844&\textbf{0.650}\\
\hline
BART&0.973&\textbf{0.816}&\textbf{0.836}&0.501&0.979&\textbf{0.874}&\textbf{0.895}&0.549\\
\hline
T5&\textbf{0.981}&0.764&0.717&0.570&\textbf{0.988}&0.870&0.842&0.602\\
\hline
    \end{tabular}
    \caption{Joint Prediction of Source, Relation and Target Prediction from Claim Sentence and Inconsistent Context Span using Multi-Task Setting}
    \label{tab:jointSRTContextSpan}
\end{table}

\noindent\textbf{Inconsistency Type and Inconsistent
Claim Component Prediction}: Tables~\ref{tab:type} and~\ref{tab:claimComponent} show the results for the inconsistency type and inconsistent
claim component prediction. Note that the two problems are 5-class and 6-class classification respectively. We observe that joint multi-task model outperforms the other two methods. Also, DeBERTa is the best model across all settings. For this best model, the F1 scores for the inconsistency types are as follows: Taxonomic Relations (0.92), Negation (0.86), Set Based (0.65), Gradable (0.78) and Simple (0.81). 
% Tried predicting mismatch span, mistype and mislocation together jointly but that did not give gains.
% Tried predicting SRT, mismatch span, mistype and mislocation together jointly but that did not give gains.

\begin{table}[!t]
    \centering
        \scriptsize
    \begin{tabular}{|l|c|c|c|c|c|c|}
\hline
Model&\multicolumn{3}{c|}{Accuracy}&\multicolumn{3}{c|}{Weighted F1}\\
\hline
&Individual&Two-step&Multi-task&Individual&Two-step&Multi-task\\
\hline
\hline
BERT&0.84&0.84&0.84&0.86&0.86&0.86\\
\hline
RoBERTa&0.85&0.85&0.86&0.86&0.86&\textbf{0.87}\\
\hline
DeBERTa&0.86&0.85&\textbf{0.87}&0.86&\textbf{0.87}&\textbf{0.87}\\
\hline
BART&0.57&0.60&0.73&0.59&0.64&0.74\\
\hline
T5&0.53&0.61&0.74&0.58&0.66&0.74\\
\hline
    \end{tabular}
    \caption{Inconsistency Type Prediction}
    \label{tab:type}
\end{table}

\begin{table}[!t]
    \centering
        \scriptsize
    \begin{tabular}{|l|c|c|c|c|}
\hline
Model&\multicolumn{2}{c|}{Accuracy}&\multicolumn{2}{c|}{Weighted F1}\\
\hline
&Individual&Multi-task&Individual&Multi-task\\
\hline
\hline
BERT&0.83&0.88&0.83&0.88\\
\hline
RoBERTa&0.85&\textbf{0.89}&0.85&\textbf{0.89}\\
\hline
DeBERTa&0.88&\textbf{0.89}&\textbf{0.89}&\textbf{0.89}\\
\hline
BART&0.80&0.75&0.81&0.76\\
\hline
T5&0.81&0.75&0.81&0.75\\
\hline
    \end{tabular}
    \caption{Inconsistent Claim Component Prediction (6-class classification)}
    \label{tab:claimComponent}
\end{table}

\begin{table}[!b]
    \centering
        \scriptsize
    \begin{tabular}{|l|c|c|c|c|}
\hline
Model&\multicolumn{2}{c|}{Accuracy}&\multicolumn{2}{c|}{Weighted F1}\\
\hline
&Individual&Individual Embedding&Individual&Individual Embedding\\
\hline
\hline
BERT&0.82&0.84&0.78&0.84\\
\hline
RoBERTa&0.83&0.86&0.80&0.85\\
\hline
DeBERTa&\textbf{0.85}&\textbf{0.87}&\textbf{0.81}&\textbf{0.86}\\
\hline
BART&0.73&-&0.71&-\\
\hline
T5&0.74&-&0.73&-\\
\hline
    \end{tabular}
    \caption{Coarse Inconsistent Entity Type Prediction. Note that embedding based methods don't work with NLG models.}
    \label{tab:coarseEntityType}
\end{table}

\noindent\textbf{Inconsistent Entity Type Prediction}: Tables~\ref{tab:coarseEntityType} and~\ref{tab:fineGrainedEntityType} show accuracy and weighted F1 for coarse and fine-grained inconsistent entity type prediction respectively. We make the following observations from these tables: (1) DeBERTa outperforms all other models for both the predictions. (2) For coarse inconsistent entity type prediction, the embedding based approach works better than the typical classification approach. This is because there are rich semantics in the entity class names that are effectively leveraged by the embedding based approach. (3) For fine-grained  inconsistent entity type prediction, two-step method is better than individual method both with and without embeddings. (4) The two-step mix method where we use embeddings based method to predict coarse inconsistent entity type and then usual 60-class classification for fine-grained types performs the best.

\begin{table}[!t]
    \centering
        \scriptsize
    \begin{tabular}{|l|c|c|c|c|c|}
\hline
Model&Individual&Two-step&Individual Embedding&Two-step Embedding&Two-step Mix\\
\hline
\hline
BERT&0.65/0.59&0.74/0.71&0.64/0.62&0.72/0.70&0.75/0.71\\
\hline
RoBERTa&0.69/0.65&0.75/0.73&0.72/0.68&\textbf{0.76}/\textbf{0.73}&0.76/0.75\\
\hline
DeBERTa&\textbf{0.70}/\textbf{0.67}&\textbf{0.77}/\textbf{0.74}&\textbf{0.73}/\textbf{0.70}&\textbf{0.76}/\textbf{0.73}&\textbf{0.78}/\textbf{0.76}\\
\hline
BART&0.50/0.44&0.64/0.59&-&-&-\\
\hline
T5&0.56/0.48&0.67/0.62&-&-&-\\
\hline
    \end{tabular}
    \caption{Accuracy/Weighted F1 for Fine-grained Inconsistent Entity Type Prediction. Note that embedding based methods do not work with NLG models.}
    \label{tab:fineGrainedEntityType}
\end{table}

\noindent\textbf{Qualitative Analysis}
To further understand where our model goes wrong, we show the confusion matrix for inconsistency type prediction for our best model in Table~\ref{tab:confusionMatrix}. We observe that the model labels many set-based examples as `taxonomic relations' leading to poor F1 for the set-based class. In general most of the confusion is between `taxonomic relations' and other classes.

\begin{table}[!t]
\centering
\scriptsize
\begin{tabular}{|l|l|c|c|c|c|c|}
\hline
\multicolumn{2}{|c|}{}&\multicolumn{5}{|c|}{Predicted}\\
\cline{3-7}
\multicolumn{2}{|c|}{}&Taxonomic Relations&Negation&Set Based&Gradable&Simple\\
\hline
\hline
\multirow{5}{*}{\rotatebox{90}{Actual}}&Taxonomic Relations&\textbf{456}&16&4&17&9\\
\cline{2-7}
&Negation&11&\textbf{123}&3&0&4\\
\cline{2-7}
&Set Based&17&4&\textbf{22}&1&1\\
\cline{2-7}
&Gradable&16&1&2&\textbf{51}&0\\
\cline{2-7}
&Simple&6&2&2&2&\textbf{36}\\
\hline
\end{tabular}
\caption{Confusion matrix for inconsistency type prediction. We observe a high correlation between actual and predicted values, indicating our model is effective.}
\label{tab:confusionMatrix}
\end{table}

% Due to lack of space, we do not put confusion matrices for other tasks. 
Amongst the coarse entity types, we found the F1 to be highest for time, action, quantity, nationality and geography entity types, and lowest for animal, relationship, gender, sentiment and technology entity types.

Further, for inconsistency spans in the context, we observe that the average length of accurate predictions (3.16) is much smaller than inaccurate predictions (8.54), comparing the lengths of ground truth spans. Further, for inaccurate predictions, we observe that as the length of the inconsistency span increases, the coverage of ground truth tokens by the predicted tokens, decreases on an average. Further, we categorized inaccurate span predictions into 4 buckets (additive, reordered, changed and subtractive). Additive implies more terms compared to ground truth, reordered means same terms but reordered, changed means some new terms were generated by the model, and subtractive means misses out on terms compared to ground truth. We found that $\sim$91 were of subtractive type, indicating that our inconsistency span predictor model is too terse and can be improved by reducing sampling probability for end of sequence token. 

% \textcolor{red}{Error Analysis}

% \textcolor{red}{Case studies}
% Examples where our model works but baselines fail.

\noindent\textbf{Hyper-parameters for Reproducibility}: The experiments were run on a machine with four GEFORCE RTX 2080 Ti GPUs. We used a batch size of 16 and the AdamW optimizer~\cite{loshchilov2017decoupled} and trained for 5 epochs for all models. We used the following models: bert-base-uncased, roberta-base, microsoft/deberta-base, facebook/bart-base, and t5-small. Learning rate was set to 1e-4 for BART and T5, and to 1e-5 for other models. More details are available in the code\footref{datafootnote}.

% \begin{itemize}
%     \item \textbf{RQ1: }Does the model performance change on datasets with smaller size? Will it help to get more annotated data?
%     \item \textbf{RQ2: }We also wanted to check if in the pipeline model, predicting relation/target before predicting the other two spans and mismatch.
%     \item \textbf{RQ3: }                                 
%                   Check the performance of the model for the mislocation class by removing the head/modifier tags.
%     \item \textbf{RQ4: }Check how the model is performing for detecting the correct entity type when the mismatching spans have named entities.
% \end{itemize}

% \subsection{Research Question Results}
\section{Conclusion and Future Work}
\label{sec:conclusion}
In this paper, we investigated the problem of detecting and explaining types of factual inconsistencies in text. We contributed a new dataset, \datasetName{}, with $\sim$8K samples with detailed inconsistency labels for (claim, context) pairs. We experimented with multiple natural language understanding and generation models towards the problem. We found that a pipeline of four models which predict inconsistency spans in claim and context followed by inconsistency type prediction and finally inconsistent entity type prediction works the best. Also, we observed that DeBERTa led to the best results. In the future, we plan to extend this work to multi-lingual scenarios. We also plan to extend this work to perform inconsistency detection and localization across multiple sentences given a paragraph.  
 % \section{Future Work}
% \label{sec:future}
% also list issues relating to dataset and models

\newpage
\section{Ethical Statement}
In this work, we derived a dataset from FEVER dataset\footnote{\url{https://fever.ai/dataset/fever.html}}. Data annotations in FEVER incorporate material from Wikipedia, which is licensed pursuant to the Wikipedia Copyright Policy. These annotations are made available under the license terms described on the applicable Wikipedia article pages, or, where Wikipedia license terms are unavailable, under the Creative Commons Attribution-ShareAlike License (version 3.0), available at this link: \url{http://creative commons.org/licenses/by-sa/3.0/}. Thus, we made use of the dataset in accordance with its appropriate usage terms.

The \datasetName{} dataset does not contain any personally identifiable information. Details of the manual annotations are explained in Section~\ref{sec:dataset} as well as in annotationGuidelines.pdf at \url{https://github.com/blitzprecision/FICLE}. 

% \section{Limitations}
% This work was done on English data only. Inconsistency classification is important for documents written in other languages as well. We plan to explore this with multi-lingual models like XLM-R, InfoXLM, DeltaLM, mBART, mT5, etc. as part of future work.

\bibliographystyle{splncs04}
\bibliography{main}
\end{document}